\documentclass[sigconf,nonacm]{acmart}
\usepackage{multirow}
\usepackage{subfloat}
\usepackage{subcaption}
\usepackage{algorithm,algpseudocode}
\usepackage{adjustbox}
\DeclareMathOperator*{\argmax}{argmax}
\usepackage{makecell}
\AtBeginDocument{%
  \providecommand\BibTeX{{%
    \normalfont B\kern-0.5em{\scshape i\kern-0.25em b}\kern-0.8em\TeX}}}

\begin{document}
% SeLaB: Semantic Labeling with BERT
\title{Semantic Labeling Using a Deep Contextualized Language Model}

\author{Mohamed Trabelsi }
\email{mot218@lehigh.edu}
\affiliation{%
  \institution{Lehigh University}
  \streetaddress{113 Research Drive (Building C)}
  \city{Bethlehem}
  \state{PA}
  \country{USA}
  \postcode{18015}
  \authornote{Mohamed Trabelsi's work was done during a summer internship in Nokia Bell Labs.}
}

\author{Jin Cao}
\email{jin.cao@nokia-bell-labs.com}
\affiliation{%
  \institution{Nokia Bell Labs}
  %\streetaddress{113 Research Drive (Building C)}
  \city{Murray Hill}
  \state{NJ}
  \country{USA}
  %\postcode{18015}
}

\author{Jeff Heflin}
\email{heflin@cse.lehigh.edu}
\affiliation{%
  \institution{Lehigh University}
  \streetaddress{113 Research Drive (Building C)}
  \city{Bethlehem}
  \state{PA}
  \country{USA}
  \postcode{18015}
}

\begin{abstract}
Generating schema labels automatically for column values of data tables has many data science applications such as schema matching, and data discovery and linking.
%The motivations behind the task are multi-fold. 
For example, automatically extracted tables with missing headers can be filled by the predicted schema labels which significantly minimizes human effort. Furthermore,  the predicted labels can reduce the %inconsistency in the naming of data columns
impact of inconsistent names across multiple data tables.
Understanding the connection between column values and contextual information is an important yet neglected aspect as previously proposed methods treat each column independently. In this paper, we propose a context-aware %schema label generation 
semantic labeling method using both the column values and context. Our new method is based on a new setting for %generating schema labels
semantic labeling, where we sequentially predict labels for an input table with missing headers. We incorporate both the values and context of each data column using the pre-trained contextualized language model, BERT, that has achieved significant improvements in multiple natural language processing tasks. To
our knowledge, we are the first to successfully apply BERT to solve the semantic labeling task.
We evaluate our approach using two real-world datasets from different domains, and we demonstrate substantial improvements
in terms of evaluation metrics over state-of-the-art feature-based methods. 

\iffalse
Generating schema labels for column values of data table is crucial for multiple data science tasks including schema matching, and data discovery and linking. The motivations behind the task are multi-fold. First, a lot of automatically extracted tables have missing headers that can be filled by the predicted schema labels which significantly minimizes human effort. Second, the predicted labels can reduce the %inconsistency in the naming of data columns 
impact of inconsistent names across multiple data tables.
Understanding the connection between column values and contextual information is an important yet neglected aspect as previously proposed methods treat each column independently. In this paper, we propose a context-aware %schema label generation 
semantic labeling method using both the column values and context. Our new method is based on a new setting for %generating schema labels
semantic labeling, where we sequentially predict labels for an input table with missing headers. We incorporate both the values and context of each data column using the pre-trained contextualized language model, BERT, that has achieved significant improvements in multiple natural language processing tasks.  
We evaluate our approach using two real-world datasets, and  demonstrate substantial improvements
in terms of evaluation metrics over state-of-the-art feature-based methods.
\fi
\end{abstract}


\maketitle
\section{Introduction}
In this era of Big Data, various datasets are publicly available for users to explore vast amounts of information in multiple fields. 
%With the large increase in the number of datasets, integrating heterogeneous datasets becomes more challenging.
Among all types of publicly available datasets, data tables represent the most prevalent form of data. A data table has multiple rows and columns. %that are %statistically independent observations, and multiple columns. 
Each column can be seen as a variable described by a schema label in order to distinguish between the variables.
Some of these data tables are pre-processed, for example, those found 
in repositories such as UCI machine learning repository\footnote{http://archive.ics.uci.edu/ml/index.php},
kaggle\footnote{https://www.kaggle.com/}, and 
OpenML \cite{openml}. Governments also store their data in a tabular format, like data.gov\footnote{https://www.data.gov/}.
In addition to that, vast amounts of information that are related to scientific, political, and cultural topics, are found on the Web. % (for example Wikipedia).
Some others require extraction such as those HTML tables that are embedded in web pages and spreadsheet files. For pre-processed tables, data providers describe their data tables using metadata, and header's names that semantically describe columns values. However, a large number of data tables do not follow metadata standards and naming standards for schema labels which leads to less informative column's labels \cite{kang2003,jaiswal2013,zhiyu,mueller2019}. For example, the date of birth of a person is saved as \textit{Date of Birth} in some data tables, and \textit{DOB} in others. %Given an unseen data table, our objective is to generate a schema label for each column from a set of labels. %Generating schema labels is a challenging Big data problem due to both the volume (billions of tables) and variety (often, the schemas of tables are very different).
On the other hand, the extracted tables can have missing or wrong headers names, as a result of automatic table extraction.  

Given an unseen data table, our objective is to generate a schema label for each column from a set of labels. Schema labels of datasets are used in multiple tasks such as data discovery \cite{castro2018,castro2_2018}, schema matching \cite{rahm2001,zapilko2012} and data preparation and analysis \cite{raman2001}.
Existing methods generate schema labels solely on the basis of their content or data values, and thus ignore the contextual information of each column when predicting schema labels. For example, both columns with labels \textit{nationality} and \textit{location}  can contain data values from the class \textit{country}, but the context of these two columns within the data table, such as other columns in the data table, has the potential to solve the ambiguity when inferring the label. In addition to that, prior approaches \cite{zhiyu,sherlock,mueller2019} define a set of hand-crafted features for each column using data values. These methods require a feature engineering phase to define, extract and validate a predefined set of specific features for the schema label generation task. The features are then used to train a traditional supervised machine learning algorithm or deep neural network architectures in order
to predict the label of previously-unseen data values. %So, in prior methods, 
In conclusion, many prior methods decouple the feature extraction and model building steps and
 require significant human effort to validate both phases. Other approaches \cite{Chen2019LearningSA,chenAAAI19} integrate external knowledge bases (KB) to predict semantic labels, with a strong assumption that the vocabulary of data tables match the KB entities.
\iffalse
\begin{figure}[!t]
\centerline {\includegraphics[width=3.5in]{model.png}}
\vspace{-.2in}
\vskip +0.1 true in \caption{Top-k accuracy results for WikiTables collection}
\label{arch}
\end{figure}
\fi

\begin{figure*}
\centering
\begin{subfigure}{.5\textwidth}
  \centering
  \includegraphics[width=1\linewidth]{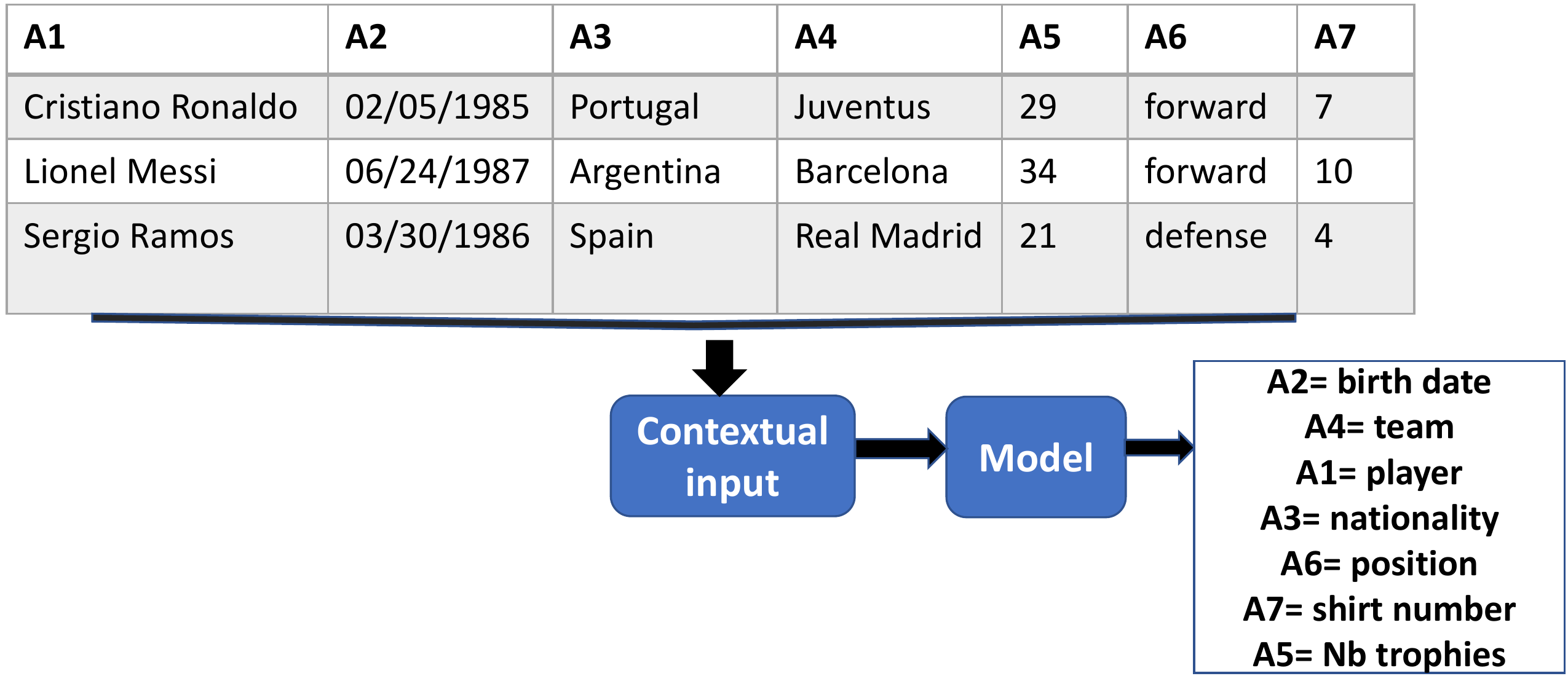}
  \caption{%Example of table that is related to Soccer
  Predicting headers for a soccer players table with true labels \textit{player}, \textit{birth date}, \textit{nationality}, \textit{team}, \textit{Nb trophies}, \textit{position},  \textit{shirt number}}
  \label{fig:sub1}
\end{subfigure}%
\begin{subfigure}{.5\textwidth}
  \centering
  \includegraphics[width=1\linewidth]{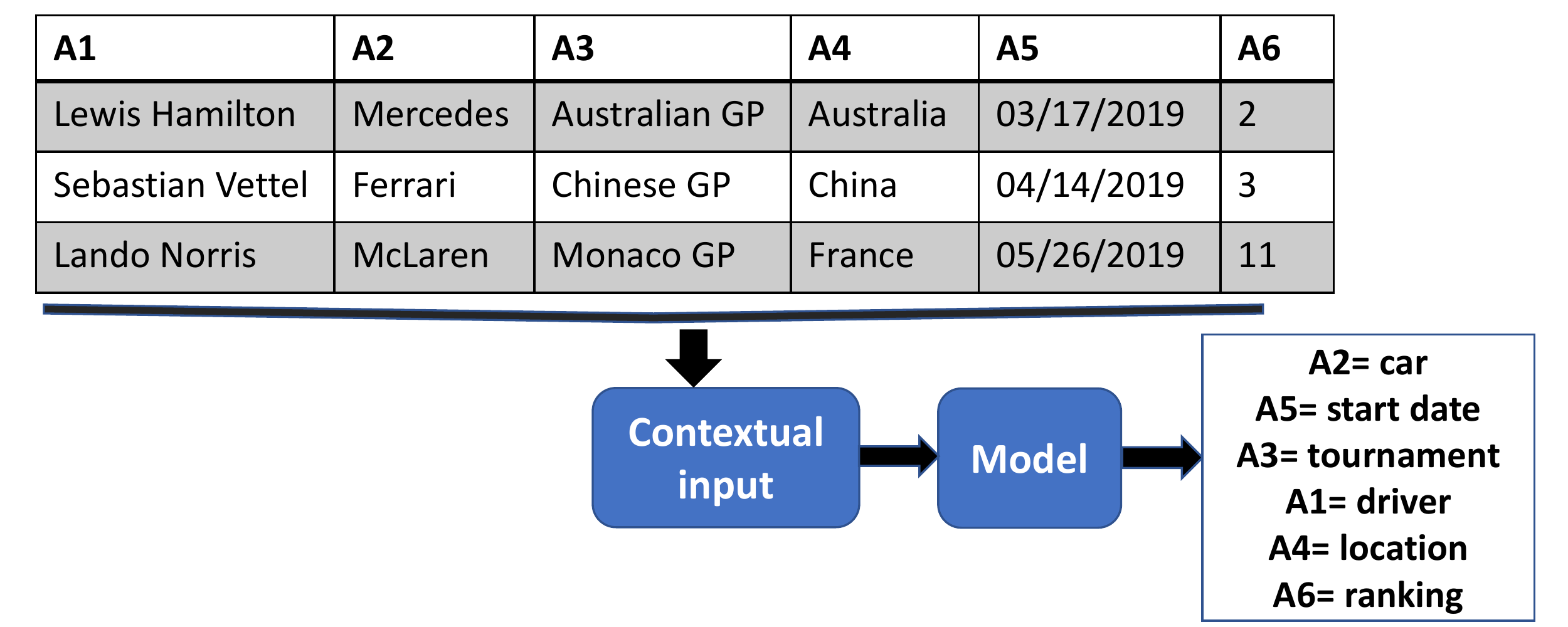}
  \caption{% Example of table that is related to Formula 1
  Predicting headers for a Formula 1 table with true labels \textit{ driver}, \textit{car}, \textit{tournament}, \textit{location}, \textit{start date}, \textit{ranking}}
  %\vspace*{-1.5mm}
  \label{fig:sub2}
\end{subfigure}
%\vspace*{-3.25mm}
\caption{The general framework of our method. The contextual input block extracts a contextual representation for each column using data values from all columns in a given table. The model block maps the contextual input of each column into a probability distribution to predict the schema label of each column. The contextual input solves ambiguity of predictions. For example, $A_3$ from Table (a), and $A_4$ from Table (b) have data values related to class \textit{country}. The context of $A_3$ using other headers, such as \textit{player}, \textit{team}, \textit{position}, etc, and the context of $A_4$ that has \textit{driver}, \textit{car}, \textit{tournament}, etc, can guide the model to predict \textit{nationality} and \textit{location} for $A_3$ (Table (a)) and $A_4$ (Table (b)), respectively. }
\label{framework}
\end{figure*}

\iffalse
%\Comment{
{\color{red}{I think we should also say that by incorporating the context, we are able to express the labels in a richer set of vocabulary unlike the limited terms that are used for describing the semantic types.}}
%}
\fi
In order to overcome the limitations of prior methods, we propose a new {\em context-aware} semantic labeling method that incorporates both data values and column's context in order to infer the label. Our method presents a new setting for generating schema labels in which the input is a table with missing schema labels or headers, instead of the traditional setting that treats each column separately. The overview of the framework, that is used in our method, is described in Figure \ref{framework}. Given a previously-unseen table with missing headers, we sequentially predict schema labels, and incorporate the already-predicted labels as context for next header prediction within the same table. 

Deep contextualized language models, like BERT \cite{Devlin2019BERTPO} and RoBERTa \cite{Liu2019RoBERTaAR}, have been recently proposed to solve multiple natural language understanding \cite{wang-etal-2018-glue,liu-etal-2019-multi} and information retrieval \cite{Yilmaz2019ApplyingBT,sakata2019,Chen2020TableSU} tasks. Different from traditional word embeddings, the pre-trained neural language models are contextual with the representation of a token is a function of the entire sentence. This is mainly achieved by the use of a self-attention structure called transformer \cite{transformer}.
Here, we integrate BERT into our proposed method, denoted by SeLaB (Semantic Labeling with BERT), to solve the schema generation task. We train a single BERT model that makes an initial prediction for the column's label using only data values, and then updates its prediction by incorporating both data values and predicted contexts of the column. SeLaB is trained end-to-end for feature extraction and model building, which reduces the significant human effort that is needed in prior methods, and gives the model the ability to capture specific features that are better than the hand-crafted ones for semantic labeling. In addition to that, by incorporating the context, we are able to predict labels in a richer and more fine-grained set of vocabulary unlike the limited classes that are used to describe the semantic labels. SeLaB doesn't assume that the column values match an existing KB, and therefore SeLaB generalizes to table collections from multiple domains. 

In summary, we make the following contributions: 
\iffalse
(1) We propose a new context-aware semantic labeling approach. Our method presents a new setting in which the input to our model is a data table with missing headers, and we sequentially generate schema labels for each %tabular data
data table. (2) We demonstrate %the usefulness of an attribute's predicted contexts that is incorporated into BERT to infer a context-aware schema label.
that by incorporating the predicted contexts of an attribute into the model, we can more accurately infer its context-aware schema label.
(3) We incorporate data values and predicted contexts using BERT, which is trained end-to-end for feature extraction and label prediction. This reduces human effort in the semantic labeling. (4) We experiment on two datasets (public and internal data table corpus), and demonstrate that our new method outperforms the state-of-the-art baselines, and generalizes to table collections from multiple domains.
\fi
%\iffalse
\begin{itemize}
    \item We propose a new context-aware semantic labeling approach. Our method presents a new setting in which the input to our model is a data table with missing headers, and we sequentially generate schema labels for each %tabular data
data table.
    \item We demonstrate %the usefulness of an attribute's predicted contexts that is incorporated into BERT to infer a context-aware schema label.
that by incorporating the predicted contexts of an attribute into the model, we can more accurately infer its context-aware schema label.
    \item We are the first to integrate BERT into the semantic labeling task. In particular, we incorporate data values and predicted contexts using BERT, which is trained end-to-end for feature extraction and label prediction. This reduces human effort in the semantic labeling.
    
    \item We experiment on two datasets (public and internal data table corpus), and demonstrate that our new method outperforms the state-of-the-art baselines, and generalizes to table collections from multiple domains.
\end{itemize}
%\fi
\section{Related work}
\subsection{Semantic Labeling}
Existing approaches \cite{pmlr-v70-valera17a,ruemmele2018,sherlock}  in the semantic labeling consist of classifying data values into a predefined set or categories
known as semantic labels. These approaches rely on a multiclass classification setup where the labels are manually defined and curated. Commercial tools, like tableau\footnote{https://www.tableau.com/} and
Trifacta\footnote{https://www.trifacta.com/}, are proposed for semantic type detection. Only a limited set of semantic types are predicted using these tools.
%Only a limited set of semantic types are predicted using commercial tools.
Hulsebos et al.~\cite{sherlock} extend the set of semantic types by considering 275 DBpedia~\cite{dbpedia} properties. These manually defined concepts, like \textit{Birth place}, \textit{Continent}, and \textit{Product}, represent the semantic types that are frequently found in datasets.
In order to infer the semantic type of a column using data values, the authors define four categories of features which are: global statistics, character distributions, pretrained word embedding, and trained paragraph embedding. 
Each feature category has different performance and noise level, so that the authors propose a multi-input neural networks model, instead of simply concatenating all features, and feeding the resulting feature vector to a single-input neural network. The multi-input neural networks model is composed of multiple identical subnetworks without weights sharing. Each subnetwork consists of two fully connected hidden layers with batch normalization, rectified linear unit (ReLU) activation functions, and dropout.
Knowledge Base based methods \cite{Chen2019LearningSA,chenAAAI19} integrate DBpedia \cite{dbpedia} to predict semantic labels, where entities from DBpedia that match all the column cells are used as additional information for a given column values.

Semantic types use a limited set of vocabulary, and can restrict the number of categories that can be considered when inferring the label of a given column. In practice, the predefined set of semantic types may not apply for new datasets. Chen et al. \cite{zhiyu} proposed a schema label generation task, in which the objective is to infer the {\em schema label}, and not only the semantic type. This setting can be seen as a multiclass classification task, where each column's label in the training set represents a possible semantic label.  Generating schema labels is more challenging as the number of possible labels is large compared to the predefined set of semantic types. The authors extract hand-crafted features from data values to predict schema labels. The set of features include content and unique content ratio \cite{content_ratio}, and the content histogram which is a 20-dimensional vector extracted using fast Fourier transform (FFT). Random forest classifier is used to predict schema labels from the curated features.

Schema matching is related to semantic type detection where the objective is to find correspondence between attributes in different schemas. Existing data on the Web, such as WebTables \cite{WebTables}, and knowledge bases, such as DBPedia \cite{dbpedia} and Freebase \cite{Freebase}, are used in schema matching. Syed et al. \cite{syed2010} use headers and data values to predict the class of a column in the target ontology or
knowledge base. The data values provide additional information that can disambiguate between the possible candidates. Limaye et al. \cite{limaye2010} associate one or more types from YAGO \cite{Yago} with each attribute or column in the table using a probabilistic graphical model. Another probabilistic approach, that is based on the maximum likelihood hypothesis, is introduced by Venetis et al. \cite{venetis2011}. The best label is chosen to maximize the probability of the values given the class label for a given column. The authors showed that class labels that are automatically extracted from the web provide more coverage for column's labeling than using manually created knowledge bases like YAGO \cite{Yago} and Freebase \cite{Freebase}. 

Matching functions are used to infer the correct semantic labels for data values. Pham et al. \cite{Pham2016SemanticLA} solve the semantic labeling as a combination of many binary classification problems. After extracting similarity metrics features from a pair of attributes, each feature vector is given a True/False label, where True means that the attributes have the same semantic type, and False indicates %the opposite. 
that the attributes are not sharing the same semantic type.
Logistic Regression and Random Forests are used to predict the matching score. For the similarity metrics features, the authors investigated multiple metrics, such as Jaccard similarity \cite{Manning2008},
cosine similarity of the product of term
frequency (TF) and inverse document frequency (IDF), known as 
TF-IDF \cite{Manning2008}
, Kolmogorov-Smirnov test (KS test) \cite{lehmann2005testing}, and Mann-Whitney test (MW test) \cite{lehmann2005testing}.
%, KS test and MW test \cite{lehmann2005testing}.  
Mueller and Smola \cite{mueller2019} proposed a neural network embeddings for data values to predict the matching score of two sets of data values. The matching score is estimated using the distance between the embeddings of two sets of data values. The score is adjusted using the output of another neural network to distinguish two columns that are different but their data values are identically distributed.

Semantic types prediction is formalized as a ranking problem in the approach proposed by Ramnandan et al. \cite{Ramnandan2015AssigningSL}. Training data values are considered as documents, and the previously-unseen data values are considered as queries. So, in the prediction phase, the objective is to extract the top $k$ candidate semantic labels for the new data values by ranking semantic labels in decreasing order using cosine similarity between query feature and every document feature in training data. The authors distinguished between textual and numeric data. For textual data, the feature vector is a weighted bag of words with TF-IDF. For numerical data, the authors used a statistical hypothesis testing to analyze the distribution of numerical data values that corresponds to a given semantic label. The statistical hypothesis test is performed between each sample in the training data and the testing sample. The returned p-values are then ranked in descending order to predict the top $k$ candidate semantic labels for the testing data values. 

Our proposed method is based on the multiclass classification setting because the schema labels are easily collected from data table corpus, unlike matching based  strategy that requires additional human effort to define pairs of attributes that have similar %labels. 
semantic type.

\subsection{BERT}

BERT \cite{Devlin2019BERTPO} is a deep contextualized language model that contains multiple layers of transformer \cite{transformer} blocks. Each transformer
block has a multi-head self-attention structure followed by a feed-forward network, and it outputs contextualized embeddings or hidden states 
for each token in the input. BERT is trained on unlabeled data over two pre-training tasks which are the masked language model, and next sentence prediction. Then, BERT can be used for downstream tasks on single text or text pairs using special tokens ([SEP] and [CLS]) that are added into the input. For single text classification, BERT special tokens, [CLS] and [SEP], are added to the beginning and the end of the input
sequence, respectively. For applications that involve text pairs, BERT encodes the text pairs with bidirectional cross attention between the two sentences. In this case, the text pair is concatenated using [SEP], and then treated by BERT as a single text. 

The sentence pair classification setting is used to solve multiple tasks in information retrieval including document retrieval \cite{dai2019,Nogueira2019MultiStageDR,Yang2019SimpleAO}, frequently asked question retrieval \cite{sakata2019}, passage re-ranking \cite{Nogueira2019PassageRW}, and table retrieval \cite{Chen2020TableSU}. The single sentence setting is used for text classification \cite{sun2019,yu2019}. BERT takes the final hidden state $\mathbf{h_{\theta}}$ of the first token [CLS] as the representation of the whole input sequence, where $\theta$ denotes the parameters of BERT.
Then, a simple softmax layer, with parameters $W$, is added on top of BERT to predict the probability of a given label $l$: $p(l \mid \mathbf{h_{\theta}})=\operatorname{softmax}(W \mathbf{h_{\theta}})$.
\iffalse
\begin{equation}
p(l \mid \mathbf{h_{\theta}})=\operatorname{softmax}(W \mathbf{h_{\theta}})
\end{equation}
\fi
The parameters of BERT, denoted by $\theta$, and the softmax layer parameters $W$ are fine-tuned by maximizing the log-probability of the true label.
%\iffalse

The organization of the rest of the paper is as follows. Section 3 formalizes the semantic labeling as a multiclass classification problem; Section 4 proposes a new context-aware semantic labeling approach that is based on the deep contextualized language model, BERT; and Section 5 illustrates data table collections that are used in our approach,
and compares baselines and our algorithm in the semantic labeling task.
%\fi

\section{Problem statement}

Our goal is to generate schema labels or semantic types for tables columns using data values, and predicted contexts in order to resolve the ambiguity problem in the prediction phase. As we mentioned before, we use the multiclass classification setting to solve schema labeling. The training data consists of a table corpus $\mathcal{T}=\{T_1, T_2,\ldots,T_n\}$, with $n$ is the total number of data tables. Each table $T_k$ has a set of $m$ columns $A_1,A_2,\dots,A_m$, where each column $A_i$ has a schema label $l_i$ (column's header), and a set of data values $V_i=\{v_1,v_2,\ldots,v_r\}$, where $r$ is the number of rows in $T_k$. The set of all possible schema labels is denoted by $L$. Resolving ambiguity when predicting schema labels requires the whole table as input to the model, instead of only using independent column's values. Therefore, our setting consists of table inputs that have missing headers, and our objective is to predict schema labels for all columns of the input table. 

We denote our proposed model by $M=N\circ F$, with $F$ is the feature extractor function (Contextual input block in Figure \ref{framework}), and $N$ is the classification layer (Model block in Figure \ref{framework}). The input to $M$ is a table $T_k$ with missing schema labels, and the output of our model is a sequence of predicted schema labels $\hat{A_1},\hat{A_2},\ldots,\hat{A_m}$. Our method learns both features and model simultaneously leading to significant reduction in human's effort spent in the feature engineering phase.

\section{Context Prediction for Semantic Labeling} \label{cper}

In this section, we introduce our context-aware method for schema label generation. We formally define the contextual information of each column, which is combined with column's data values to improve the performance of semantic labeling.

\subsection{Column's context}

The set of data values $V_i$ for a given column $A_i$ in a table $T_k$ are not sufficient to have accurate schema label prediction. For example in Figure  \ref{framework}, both columns \textit{nationality} (A3 in the left table) and \textit{location} (A4 in the right table) contain values from class \textit{country}, but they refer to different labels. In this case, if we know that $A_3$ (in the left table) occurs in a table that contains \textit{player}, \textit{team}, \textit{position}, and \textit{birth date} attributes, hence it is more probable that $A_3$ is related to \textit{nationality} rather than \textit{location}. Therefore, we argue that the attributes $A_1,A_2,\ldots,A_{i-1},A_{i+1},\ldots,A_m$ provides a rich contextual information for $A_i$. However, as we explained in our setting, the input to our model is a table that has missing headers, which means that we cannot directly incorporate the context into our model. 

To solve that, we propose incorporating predicted context instead of the ground truth context. In other words, our model has two passes for predicting schema labels. During the first pass, given a table $T_k$ with missing headers, only data values are used to make initial predictions for semantic labels, denoted by $A_1',A_2',\ldots,A_m'$. The initial predictions are context-free, as they only capture data values. For the second pass, we incorporate both data values $V_i$, and the predicted context $A_1',A_2',\ldots,A_{i-1}',A_{i+1}',\ldots,A_m'$ of $A_i$ to make the final context-aware prediction, denoted by $\hat{A_i}$.

\begin{algorithm}[!h]
\small
	\caption{Training phase} 
	\label{training}
	
	\begin{algorithmic}[1]
	\State %\hspace*{\algorithmicindent}
	\textbf{Input}: tables collection $\mathcal{T}=$\{$T_1, T_2,\ldots,T_n$\}, set of %headers 
	labels $L$. %and $(\theta_{old},W_{old})$ initial parameters of $M$.
		%\For {$epoch=1,2,\ldots,E$}
			\For {$T_k$ in $\mathcal{T}$}
			    \State the schema labels of $T_k$, $l_1,l_2,\ldots,l_m$, are available
    			\State \% First phase: Values based prediction phase
    			    \For {$[A_i,V_i]$ in $T_k$}
        				\State input to BERT for $A_i$: $I_1(A_i)=[CLS]+V_i+[SEP]+ %\hspace*{1.5cm}
        				[SEP]$
        				%\State compute values based prediction: 
        				%= probability \hspace*{1.5cm} distribution over all headers $L$, $p_i'=M(I_1(A_i))$ 
        				\State compute values based 
        				probability, $p_i'=M(I_1(A_i))$
        				\State $A_i'= \argmax_{l \in L} p_i'[l]$
    				\EndFor
    			\State \% Second phase: Compute contexts of each attribute %using \hspace*{1.0cm} first phase predictions
    			%\State $m=$ number of attributes in $T_k$
    			\For {$[A_i,\_]$ in $T_k$}
        				\State Context of $A_i$: $context(A_i)=A_1'+A_2'+\ldots+A_{i-1}'+\hspace*{1.5cm} A_{i+1}'+\ldots+A_{m}'$
        				\State avoid true label leakage in context: remove $l_i$ from \hspace*{1.5cm} $context(A_i)$  if $l_i \in context(A_i)$ 
        				\State remove duplicates from $context(A_i)$
    			\EndFor
    			\State \% Third phase: compute final predictions
    			\For {$[A_i,V_i]$ in $T_k$}
        				\State input to BERT for $A_i$: $I_2(A_i)=[CLS]+V_i+[SEP]+ \hspace*{1.5cm} context(A_i)+[SEP]$
        				%\State compute context based prediction:  
        				 %probability \hspace*{1.5cm} distribution over all headers $L$, $\hat{p_i}=M(I_2(A_i))$ 
        				 \State compute context-aware
        				 probability, $\hat{p_i}=M(I_2(A_i))$
        				\State $\hat{A_i}= \argmax_{l \in L} \hat{p_i}[l]$
    			\EndFor
    			
    			\State $loss(T_k)$=$CrossEntropy([\hat{p_1},\hat{p_2},\ldots,\hat{p_{m}}]$, %\hspace*{1.5cm}
    			$[l_1,l_2,\ldots,l_{m}])$
    			\State update $M$ parameters $(\theta,W)$  by minimizing $loss(T_k)$  %($\theta_{old}\leftarrow\theta$, $W_{old}\leftarrow W$)
			%\State $\theta_{old}\leftarrow\theta$, $W_{old}\leftarrow W$
			\EndFor
		%\EndFor
	\State %\hspace*{\algorithmicindent}
	\textbf{Output}: A Trained %$M$ that predicts header based on values and predicted contexts 
	context-aware model $M$
	\end{algorithmic} 
\end{algorithm}

\subsection{Semantic labeling with BERT (SeLaB)}

We incorporate data values and predicted contexts of a given attribute using the contextualized language model BERT. So, for our proposed model $M=N\circ F$, denoted by {\em SeLaB}, $F$ is equivalent to BERT with parameters $\theta$, as we use the hidden state of [CLS] token from the last transformer block 
to compute the embedding of the input
%a 
sentence. $N$ denotes the softmax layer
with parameters $W$ that is used to produce the probability distribution of a given sequence over all schema labels from $L$. The general form of input to $M$ for an attribute $A_i$, denoted by \textit{contextual input}, is the sequence [CLS]+$V_i$+[SEP]+$context(A_i)$+[SEP], where $context(A_i)$ is the predicted context of $A_i$. For first pass prediction, where $context(A_i)$ is missing, the input sequence form, denoted by \textit{only values}, becomes [CLS]+$V_i$+[SEP]+[SEP]. Next, we describe the training and testing phases.

\subsubsection{Training phase}

The steps of training phase are shown in Algorithm \ref{training}. The inputs to training phase are: table corpus $\mathcal{T}=\{T_1, T_2,\ldots,T_n\}$ where semantic labels $l_1,l_2,\ldots,l_m$ are available for all attributes $A_1,A_2,\dots,A_m$  of a given table $T_k \in \mathcal{T}$, set of possible semantic labels $L$, and pre-trained BERT model as a feature extractor $F$. The compact notation of table $T_k$, that is used in algorithms, is $T_k=[[A_1,V_1],[A_2,V_2],\ldots,[A_m,V_m]]$.

The training process has three phases. The first phase consists of predicting an initial label for each column using \textit{only values} input form as shown in Lines 4--9 of Algorithm \ref{training}. The output of the first phase is a sequence $A_1',A_2',\ldots,A_m'$ of initial predicted labels. During the second phase (Lines 10--15), we construct the predicted context $context(A_i)$ for each attribute $A_i$, which is the set of predicted labels $\{A_j';j \in [1,m]\setminus\{i\}\}$. In order to avoid the true label leakage in $context(A_i)$, we remove $l_i$ from $context(A_i)$  if $l_i \in context(A_i)$. We also remove duplicates from $context(A_i)$ as most of data tables contain unique headers. The final phase (Lines 16--21) computes the context-aware predictions by using \textit{contextual input} form. The output of $M$ is the probability distribution $\hat{p_i}$ over all labels in $L$, for every $A_i \in T_k$. These probability distributions are used to calculate the cross entropy loss, and to update the parameters of $M$ as indicated in Lines 22--23. In addition to incorporating the context of column for schema labeling, 
our model has the ability to accept two forms of sequence inputs (\textit{only values} and \textit{contextual input}), which significantly reduces the number of parameters compared to the case where a separate model is needed to handle each type of input sequence. 

In contrast to \cite{Ramnandan2015AssigningSL,Pham2016SemanticLA} which have a pre-processing step to distinguish between string and numerical attributes, our BERT-based feature extractor $F$ is able to process string and numerical texts by taking advantage of BERT tokenizers. In contrast to \cite{zhiyu,sherlock} where the feature extraction and model building steps are decoupled, our model $M=N\circ F$ is trained end-to-end to jointly optimize the feature extractor $F$, and the classification layer $N$. Unlike \cite{Chen2019LearningSA,chenAAAI19} that integrate external KB in semantic labeling with a strong assumption that the column's values match the KB entities, SeLaB needs only BERT embeddings that is fine-tuned on target table corpus to extract the feature of each column, and therefore generalizes to data tables from multiple domains. We train SeLaB for $E$ epochs.
\begin{algorithm}
\small
	\caption{Testing phase} 
	\label{testing}
	\begin{algorithmic}[1]
	\State %\hspace*{\algorithmicindent}
	\textbf{Input}: testing table $T_k$, set of headers labels $L$, trained $M$ model, Boolean: $unique\_headers$ and $top_k$.
		%\State $m=$ number of attributes in $T_k$
		%\State the schema labels of $T_k$, $l_1,l_2,\ldots,l_m$, are missing
		\State First phase: Values based prediction phase
		    
		\State Second phase: Compute contexts of each attribute %using first phase predictions
		
		\State \% Third phase: compute final predictions
		\State $predicted\_attributes=\emptyset$, $seen\_columns=\emptyset$
		\For {$it$ in $[1,m$]}
		    \For {$[A_i,V_i]$ in $T_k$}
		    
				\State input to BERT for $A_i$: $I_2(A_i)=[CLS]+V_i+[SEP]+\hspace*{1cm} context(A_i)+[SEP]$
        				%\State compute context based prediction:  
        				 %probability \hspace*{1cm} distribution over  all headers $L$, $\hat{p_i}=M(I_2(A_i))$ 
        				 \State compute context-aware
        				 probability, $\hat{p_i}=M(I_2(A_i))$ 
        				\State $\hat{A_i}= \argmax_{l \in L} \hat{p_i}[l]$, \hspace*{\algorithmicindent}
        				  $p^i_{max}= \max_{l \in L} \hat{p_i}[l]$
		    
		    \EndFor
		    \If{$unique\_headers$}
                \State $h,chosen\_label=UniqueHeaders (\{\hat{p_1},\hat{p_2},\ldots, \hat{p_{m-it+1}}\},$ \hspace*{1cm} $ top_k, predicted\_attributes)$
                
                \State $\hat{A_h}=chosen\_label$
            
            \Else
                \State $h= \argmax_{s \in [1,m-it+1]} p^s_{max}$
            \EndIf
                
		    \State $predicted\_attributes.append(\hat{A_h})$, \State $seen\_columns.append(h)$
		    \State $T_k.delete([A_h,V_h])$
		    \State update contexts with new predicted attribute $\hat{A_h}$:
		     replace  $A_h'$ by $\hat{A_h}$ \hspace*{0.5cm}  in  $context(A_i)$, $i \in [1,m]\setminus\{seen\_columns\}$
				
		\EndFor
			
	\State %\hspace*{\algorithmicindent}
	\textbf{Output}: A label for each header in the testing table. 
	\end{algorithmic} 
\end{algorithm}
\subsubsection{Testing phase}
The steps of the testing phase are shown in Algorithm \ref{testing}. The inputs to the testing phase are: a testing table $T_k$ that has missing headers ($l_1,l_2,\ldots,l_m$, are not available), set of possible semantic labels $L$, trained model $M$, and two parameters $unique\_headers$ and $top_k$ that we will describe later. 

The testing process has three phases. The first and second phases (Lines 2--3) are similar to the training process, where initial predictions are computed using \textit{only values} input form, and then used to produce the context of each attribute. During the third phase, the final predicted labels for the testing data table are generated sequentially as shown in Lines 4--22. For a given table $T_k$, initially all schema labels are missing , and the set of predicted attributes, denoted by $predicted\_attributes$, is empty. Given that the prediction is done sequentially, $m$ passes are needed to obtain a predicted schema label for each column in $T_k$. For the $j$-th pass, the $predicted\_attributes$ has $j-1$ labeled headers, and $m-j+1$ columns in $T_k$, denoted by $S_j$, are still missing the predicted labels. We predict the probability distribution $\hat{p_i}$, and a schema label $\hat{A_i}$ for each column $A_i \in S_j$ using our model $M$ with \textit{contextual input} sequence. The confidence of prediction for $A_i \in S_j$ is given by $p^i_{max}= \max_{l \in L} \hat{p_i}[l]$. The $unique\_headers$ is a Boolean variable that we set to \textit{True} %in case we want 
to force the unique headers constraint for a given table. When predicting duplicate headers is allowed, the column $h$, that we choose to predict from $S_j$ in the $j$-th pass, is given by $h= \argmax_{s \in [1,m-j+1]} p^s_{max}$ as shown in Lines 15--16. 

%\iffalse
\begin{algorithm}
\small
	\caption{$UniqueHeaders$} 
	\label{constraint}
	
	\begin{algorithmic}[1]
	\State %\hspace*{\algorithmicindent}
	\textbf{Input}: probability distributions $\hat{p_1},\hat{p_2},\ldots,\hat{p_{l_m}}$,  $top_k$, already predicted headers set $P_h$
	
	\State $p_i^k$: select $top_k$ probabilities of $\hat{p_i}$, $i \in [1,l_m]$ 
	\State $A_i^k$: $argmax$ of $top_k$ probabilities of $\hat{p_i}$, $i \in [1,l_m]$
	
	\Repeat
	\State $c=\argmax_{s \in [1,l_m]} p_s^k[1]$
	\State $chosen\_label=A_c^k[1]$
	%\State update $p_c^k$ and $A_c^k$ for next iteration if $chosen\_label \in  P_h$
	\State $p_c^k[1:l_m-1]=p_c^k[2:l_m]$,\hspace*{\algorithmicindent} $p_c^k[l_m]=-1$
	\State $A_c^k[1:l_m-1]=A_c^k[2:l_m]$,\hspace*{\algorithmicindent} $A_c^k[l_m]=-1$
	
	\Until{$chosen\_label \notin P_h$ or $p_i^k$ contains only $-1$, for $i \in [1,l_m]$}
	
	\If{$chosen\_label \in P_h$}
       \State return $chosen\_label$ from first pass in $repeat$ loop
    \EndIf
    
\State %\hspace*{\algorithmicindent} 
\textbf{Output}: $c$: chosen column for prediction , %predicted attribute 
$chosen\_label$
\end{algorithmic} 

\end{algorithm}
\setlength{\textfloatsep}{4pt}
%\fi

On the other hand, when unique headers constraint is required for a given data table, we propose a routine, called $UniqueHeaders$, that resolves the duplicate headers problem as shown in Algorithm \ref{constraint}. %This routine is described in Algorithm \ref{constraint}.
The inputs to this routine are: the probability distributions $\hat{p_i}$ for $A_i \in S_j$, $top_k$ which denotes the number of top confidences per attribute that are used to find the label,
and the set $predicted\_attributes$ that contains the $j-1$ semantic labels that are already assigned to $j-1$ columns of $T_k$. The objective of %Algorithm \ref{constraint}
the function %$Force\_unique\_headers\_in\_table$
$UniqueHeaders$ is to find the label $chosen\_label$ with the highest confidence value, with respect to the unique headers constraint that requires $chosen\_label \notin predicted\_attributes$. %Algorithm \ref{constraint} 
%This routine
%can be seen as a %greedy 
%\textit{breadth first search}, with each level has $m-j+1$ nodes. %, and the maximum depth of search is equal to $top_k$. 
For time complexity efficiency, we limit the depth of search by choosing $top_k<<|L|$. By limiting the depth of search, %Algorithm \ref{constraint}
$UniqueHeaders$ can produce a duplicate header. In this case, we use a heuristic that returns the label that corresponds to the maximum confidence score. %The routine, that is presented by Algorithm \ref{constraint},
$UniqueHeaders$ is called in Lines 12--14 of Algorithm \ref{testing}.

We remove the chosen column $h$ from $S_j$ to obtain $S_{j+1}$ (the columns of $T_k$ that are still missing labels after the $j$-th pass) , and we add the chosen column $h$ to $seen\_columns$ set, and the predicted label $\hat{A_h}$ to $predicted\_attributes$ set (Lines 18--20). We finish the $j$-th pass by updating $context(A_i)$ ,where $A_i \in S_{j+1}$, using the predicted label $\hat{A_h}$ from the $j$-th pass as shown in Line 21. The objective of the context update step is to replace the \textit{only values} predicted label by the \textit{contextual input} inferred schema label, as the latter is more accurate than the former. For the $j$-th pass, we select the best schema label from $m-j+1$ predicted labels. %So, for $m$ passes, the total number of predictions, that are needed to infer the schema labels of $T_k$, is $m \times (m-1)/2$.
The increase in the number of predictions is justified by the sequential nature of the testing algorithm where context is updated in each pass, and the most confident prediction is selected.

\section{Evaluation} \label{eval}
\subsection{Data collections}

\subsubsection{WikiTables}
This dataset is composed of the WikiTables corpus \cite{Bhagavatula2015TabELEL} which contains over $1.6M$ tables that are extracted from Wikipedia. Since a lot of tables have unexpected formats, we preprocess tables so that we only keep tables that have enough content with at least $3$ columns and $50$ rows. We further filter the columns whose schema labels appear less than 10 times in the table corpus, as there are not enough data tables that can be used to train the model to recognize these labels. We experiment on $15,252$ data tables, with a total number of columns equal to $82,981$. The total number of schema labels is equal to $1088$.
\subsubsection{Log Tables from Network Equipment}
Our work is motivated by the business need to automatically generate schema labels for the  data tables extracted from log files of network equipment. Network logs files contain computer-generated event records, such as
authentication attempts, process assessment calls and information output
of network equipment, and are instrumental for network performance monitoring and fault diagnosis.
%Unfortunately, making sense out of log files are extremely challenging as someone put it ”Log files are generally a mess, and the extraction of statistical information about patterns in them is one of the most common challenges in software quality control.” 
\iffalse
For such purposes, the engineers need to parse the messy log files to extract 
important metrics and %present the information that are easier to handle
prepare information for downstream analysis. However, this is in general a messy and cumbersome process as
there is no uniform standard for the log files. We are working toward a solution that can automatically parse log files into data tables and part of this automation is auto-generating the labels that are most appropriate for the table columns.
\fi

For the purpose of schema label auto-generation, we shall utilize the existing data tables that have already been collected in an internal platform used
by network care engineers from parsing log files.  
% The tables are generated from parser for specific network 
%log files developed by engineers in a crowd-sourcing manner,
In the current pipeline, engineers design a parser for each type of log files, and these parsers generate the tables.
% as engineers that can be shared across the users of this platform. 
%where users engineers generate log tables from network logs using this platform.
We have collected 329 tables from this platform with logs files
coming 
from products on  wireless equipment such as base station, Radio Access Network and Radio Network Controllers. To evaluate our methods on header prediction, 
we removed tables that have less than 10 rows and cleaned up columns that have mostly invalid values (such as NULL, empty string, or NA). The remaining set contains 248 tables. The number of rows of these tables have a very skewed distribution with quantiles being
138 (25\%), 551 (50\%) and 1954 (75\%), while the number of columns ranges from 3 to 48 columns with many tables having in the neighbourhood of 10 columns. For our purpose, we focus on 87 headers from these tables that have more than 3 instances. 

\begin{figure}
    \centering
     \includegraphics[width=0.23\textwidth]{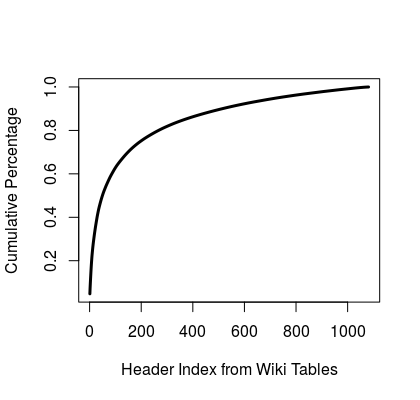}
        \includegraphics[width=0.23\textwidth]{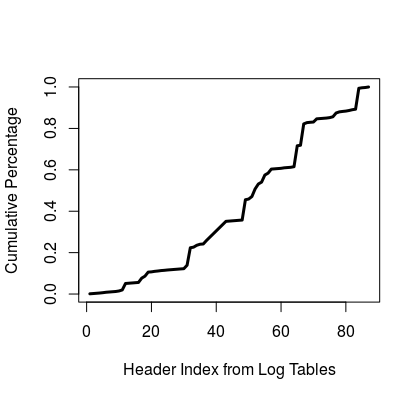} 
    %\vspace*{-4mm}
    \caption{Cumulative Frequency Distribution of Headers}
    \label{fig:cumfreq}
\end{figure}

Figure~\ref{fig:cumfreq} shows the 
cumulative frequency distribution for the headers from the two datasets, from the most to the least popular.  %It is clear that 
There is a small set of labels that are much more frequently occurring in WikiTables. One reason that the labels in log tables are more scattered is because the tables are manually collected from %a 
diverse %set of 
products as we would like to understand the performance of our algorithm in various situations.

\subsection{Baselines}	
We compare the performance of our proposed model with feature-based baselines \cite{zhiyu,sherlock}, and a variation of our model where only data values (without context) are used. We describe the five categories of features that are extracted from the data values of each column. There are five categories of features as shown in Table~\ref{categories}. 
Four categories are previously used in feature-based methods: global statistics \cite{zhiyu,sherlock}, character distributions \cite{zhiyu,sherlock}, word embeddings \cite{sherlock}, and paragraph embeddings \cite{sherlock}. To obtain more fine-grained embedding, we also propose a character-based generative model to produce character embeddings for each value.
%\iffalse
\begin{table}[!h]
\centering
\caption{Features categories used in feature based method (total dimension equals to 2213)}
\label{categories}
\begin{tabular}{@{}lll@{}}
\toprule
 ID&Category type& Dimension  \\ \midrule
1&Global statistics& 52 \\
2 &Character distributions&960 \\
3&Character embeddings&400 \\
4&Word embeddings&401 \\
5&Paragraph embeddings& 400
 \\ \bottomrule
\end{tabular}
\end{table}
%\fi
\subsubsection{Global statistics}

We combine the global statistics from \cite{zhiyu} and \cite{sherlock} into one category that has 17 unique features with different dimensions as shown in Table \ref{gsf}.
When concatenating the global statistics features, the dimension of the resulting feature vector is 52.

%\iffalse
\begin{table*}[!h]
\caption{Global statistics features \cite{zhiyu,sherlock} with a total dimension equals to 52\label{gsf}}
\small
\centering
\begin{tabular}{ ||c|c|c|| }
\hline
ID & Length&Description\\ [0.5ex] 
 \hline
  1&1  & minimum value in column's content \\ 
  \hline
  2&1  &maximum value in column's content  \\
  \hline
  3&1  &average value of column's content  \\
  \hline
 4&1  &standard deviation value of column's content  \\
 \hline
 5&1  &percentage of numeric cells in column's content  \\
 \hline
 6& 20 & content histogram \\
 \hline
 7&  1&number of values in a column  \\
 \hline
 8&  1& column entropy  \\
 \hline
 9&  1& percentage of unique content   \\
 \hline
 10&  1& percentage of values with numeric characters  \\
 \hline
 11&  1& percentage of values with alphabetical characters   \\
 \hline
 12&  2&\makecell{mean and standard deviation of the number of numerical characters in column's content}   \\
 \hline
 13& 2 &\makecell{mean and standard deviation of the number of alphabetical characters in column's content}  \\
 \hline
 14&2  & \makecell{mean and standard deviation of the number of  special characters in column's content} \\
 \hline
 15&2  & \makecell{mean and standard deviation of the number of words in column's content}  \\
 \hline
 16&4  &\makecell{None values statistics: count, percentage, has some None values(Boolean), has Only None values (Boolean)}  \\
 
 \hline
  17&10  &\makecell{length of values statistics: any nonzero length (Boolean), all nonzero length (Boolean), sum, min, max,variance,\\ median, mode, kurtosis, skewness} \\ [1ex] 
 \hline
\end{tabular}
\end{table*}
%\fi

\subsubsection{Character distributions}
The distribution of characters is computed for each column using 96 ASCII-printable characters. For each character, 10 statistics (any, all, mean, variance, min, max, median, sum,
kurtosis, skewness) are calculated based on the count of each character in a value from the set of data values for the input column. Concatenating character distribution for all characters results in a feature vector with dimension 960.

\subsubsection{Character embeddings}
We train a character-level language model \cite{sutskever2011} to generate values from the table corpus $\mathcal{T}$. Our generative model has two character-based LSTM layers, and it is trained on the next character generation task. Given a value $v_k$, which is a sequence of characters, our generative model produces a hidden state for each character, and we use the hidden state from the second LSTM layer of the last character as the character embedding of $v_k$. Then as in \cite{sherlock}, we compute the mean, mode, median and variance of character embeddings across all values $v_k \in V_i$ in a column $A_i$. Given that the dimension of character embedding is $100$, and we have $4$ statistics, the resulting feature vector has dimension $400$. 

\subsubsection{Word embeddings}

Pre-trained word embedding, such as Glove \cite{glove}, is used to compute embedding for each value $v_k \in V_i$. Then, as in character embedding, $4$ statistics are computed for $V_i$. The dimension of embedding is $100$, so that after computing statistics, the dimension of the concatenated feature vector equals $400$. As in \cite{sherlock}, an additional binary feature is appended to the final word embedding feature vector, and it indicates if there is at least one value from $V_i$ that belongs to Glove vocabulary.

The use of pre-trained embedding is suitable for table collections, such as WikiTables, that have common values with the vocabulary of the pre-trained embedding. This is not the case for log tables from network equipment, where the number of out of vocabulary (OOV) tokens is large, and this leads to poor performance for pre-trained word embedding. To solve this problem, we train a word embedding on the table corpus $\mathcal{T}$, where the sentences are rows and columns from $T_k \in \mathcal{T}$. Instead of using Glove, we train a fastText \cite{bojanowski2017} model to produce word embeddings. The use of character-level n-grams in fastText allows word embeddings to be created even for terms that have not been seen before, and reduces the negative effect of OOV tokens. Given that having long strings is common for log data, we use BERT tokenizer to preprocess values before training fastText embeddings.

\subsubsection{Paragraph embeddings}
Each column $A_i$ can be seen as a paragraph that contains the set of values $V_i=\{v_1,v_2,\ldots,v_r\}$. The paragraph embedding, that is based on the distributed bag of words \cite{le2014}, is trained to map each column into an embedding with dimension equals to $400$.

\subsubsection{Sherlock} 
Hulsebos et al.~\cite{sherlock} uses global statistics, character distributions, word embeddings, and paragraph embeddings with a multi-input neural networks architecture.

\subsubsection{All features} 
This baseline extends Sherlock \cite{sherlock} features by adding our character embeddings to cover three different levels of embeddings (character, word, and paragraph).

\subsubsection{BERT with only values}
This baseline can be seen as a variation of our proposed method
SeLaB, where only data values are used to predict schema labels. So, for training, the first phase predictions are used to update the parameters of the model. For testing, the first phase predictions are used to evaluate the performance of the model. The input data values sequence to the model has \textit{only values} input form. 

We note that we do not compare to external KB based methods \cite{Chen2019LearningSA,chenAAAI19} because there is a vocabulary mismatch between log tables from network equipment and DBpedia, and an important aspect of our evaluation is to show generalization to data tables from multiple domains (not only tables from Wikipedia and Web tables).

\begin{table*}[t!]
\footnotesize
%\small
\begin{subtable}[t]{0.48\textwidth}
\begin{tabular}{@{}llllll@{}}
\toprule
Method Name & Macro-P & Macro-R & Macro-F &Micro-F&MRR  \\ \midrule
 Global statistics&7$\times10^{-3}$& $10^{-2}$  & 6$\times10^{-3}$ & 0.17 & 0.28 \\\midrule
 Character distributions&0.14  &0.10  &0.10 &0.38 &0.48  \\\midrule
 Character embeddings&0.38  &0.33  &0.33 &0.54 &0.64  \\\midrule
 Word embeddings&0.18  &0.16 &0.15 &0.47 & 0.58 \\\midrule
 Paragraph embeddings&0.23  & 0.19 & 0.19&0.43 & 0.54 \\\midrule
 Sherlock& 0.45 & 0.45 &0.42 &0.66 & 0.75 \\\midrule
 All features&0.54  & 0.47 &0.47 &0.66 &0.75  \\\midrule
\midrule
BERT (only values)&0.46  & 0.45 &0.43 & 0.64&0.73  \\\midrule
SeLaB w/o unique headers&\textbf{0.55}  &0.52  &0.50 & \textbf{0.72}& \textbf{0.80} \\\midrule
SeLaB&\textbf{0.55}  &\textbf{0.53}  &\textbf{0.51} &\textbf{0.72} &\textbf{0.80}  
 \\ \bottomrule
\end{tabular}
\caption{\footnotesize WikiTables}
\label{tab:table1_d}
\end{subtable}
%\bigskip 
\hspace{\fill}
\begin{subtable}[t]{0.48\textwidth}
\begin{tabular}{@{}llllll@{}}
\toprule
Method Name & Macro-P & Macro-R & Macro-F &Micro-F&MRR  \\ \midrule
 Global statistics& 0.16  & 0.16  & 0.15 & 0.54 & 0.64 \\\midrule
 Character distributions&0.40  &0.42  &0.40 &0.68 & 0.71 \\\midrule
 Character embeddings&0.44 &0.44  &0.43 &0.69 & 0.71 \\\midrule
 Word embeddings&0.35  &0.34&0.33 &0.65 & 0.69 \\\midrule
 Paragraph embeddings&0.31  &0.29  &0.29 &0.61 & 0.67 \\\midrule
 Sherlock& 0.38 & 0.38 &0.37 &0.67 & 0.73 \\\midrule
 All features&0.50  &0.49  &0.49 &0.71 & 0.72 \\\midrule
\midrule
BERT (only values)&0.48  &0.49  &0.47 &0.73 & 0.76 \\\midrule
SeLaB w/o unique headers&0.61  &0.62  &0.60 &\textbf{0.81} &\textbf{0.81}  \\\midrule
SeLaB&\textbf{0.62}  &\textbf{0.63}  &\textbf{0.61} &\textbf{0.81} &  \textbf{0.81}
 \\ \bottomrule
\end{tabular}
\caption{\footnotesize Log tables}
\label{tab:table1_a}
\end{subtable}
%\vspace*{-4mm}
\caption{ Semantic labeling results %\subref{tab:table1_a} and \subref{tab:table1_d}.
}
\label{metrics}
\end{table*}

\begin{figure*}[t!]
    \centering
    \begin{subfigure}[t]{0.5\textwidth}
        \centering
        \includegraphics[height=2.5in]{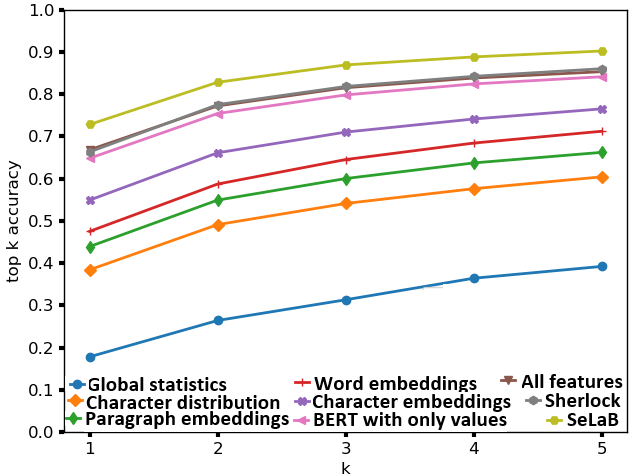}
        \caption{WikiTables collection}
    \end{subfigure}%
    ~ 
    \begin{subfigure}[t]{0.5\textwidth}
        \centering
        \includegraphics[height=2.5in]{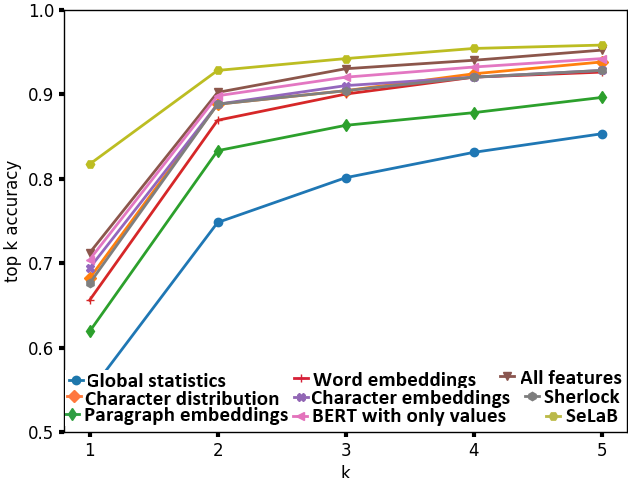}
        \caption{Log tables collection}
    \end{subfigure}
    %\vspace*{-4mm}
    \caption{Top-k accuracy results}
    \label{accuracy}
\end{figure*}

\subsection{Experimental Setup}
In our proposed model, we use the BERT-base-uncased for the feature extractor $F$. For each column, we shuffle data values and randomly select a subset of values to reduce the complexity of the model. Given that the majority of log tables have more rows than WikiTables, we randomly choose $200$ values for each column in a given log table, and $100$ values for columns from WikiTables. In general, shuffling the predicted context can reduce overfitting. For WikiTables collection, the majority of tables that have both attributes \textit{home team} and \textit{away team} (these attributes have similar data values), report \textit{home team} column before \textit{away team} column (same for \textit{birth date} which occurs before \textit{death date}) for a left-to-right sequential order. In this case, we can resolve the ambiguity of predicting \textit{home team} and \textit{away team} by keeping the sequential order of the predicted context. We train our model for 10 epochs, and we set $top_k$ of $UniqueHeaders$ routine
to $5$. 

The model is implemented using PyTorch, with Nvidia GeForce GTX 1080 Ti. We use Adam \cite{adam} optimizer for gradient descent to minimize the cross entropy loss function and update the weights of our model. 
We report results for our method and baselines using a random split of the entire table corpus, where $80\%$ of tables are used for training, and $20\%$ for testing. For the baselines, BERT with only values is also trained for $10$ epochs. For the feature-based baselines (except for Sherlock), % machine learning models, such as 
random forest %and multilayer perceptrons (MLP)
is trained for prediction. %to predict schema labels.

\subsection{Experimental results}

We evaluate the performance of SeLaB and baselines on the schema labeling task using macro-averaged and micro-averaged precision (P), recall (R) and F-score of predictions on the testing set. In the multiclass classification problem, the micro-average precision, recall and F-score are the same, so we only report the Micro-F score. We also report the Mean Reciprocal Rank (MRR) \cite{liu}, as the rank of the true class is an important measure for evaluation. In addition to that, we calculate the top-$k$ accuracy that shows the fraction of testing samples where the true label is within the top $k$ predicted confidences.

\subsubsection{Semantic labeling results}

\iffalse
\begin{figure*}[t!]
    \centering
    \begin{subfigure}[t]{0.33\textwidth}
        \centering
        \includegraphics[height=1.7in]{wiki_accuracy3.png}
        \caption{WikiTables collection}
    \end{subfigure}%
    ~ 
    \begin{subfigure}[t]{0.33\textwidth}
        \centering
        \includegraphics[height=1.7in]{log_accuracy2.png}
        \caption{Log tables collection}
    \end{subfigure}
    ~ 
    \begin{subfigure}[t]{0.33\textwidth}
        \centering
        \includegraphics[height=1.7in]{masked_headers2.png}
        \caption{SeLaB top-1 accuracy for masked headers}
    \end{subfigure}
    %\vspace*{-4mm}
    \caption{Top-k accuracy results}
    \label{accuracy}
\end{figure*}
\fi

Table \ref{metrics}(a) shows the performance of different approaches on the WikiTables collection. We show that our proposed method SeLaB outperforms the baselines for all evaluation metrics. The context, which is incorporated into our model, solves the ambiguity in predictions and leads to an increase in evaluation metrics compared to baselines that generate schema labels solely on the basis of data values. Among all the five categories of features (global statistics, character distributions, character embeddings, word embeddings, paragraph embeddings) that are used in the feature-based approaches,
our character embeddings feature achieves higher performance for all evaluation
metrics. So, a generative model with a character granularity is able to capture the distributions of data values drawn from different variables or attributes.
\iffalse
\begin{table}[]
\centering
\caption{WikiTables schema labels prediction}
\label{wikitables_results}
\begin{tabular}{@{}llllll@{}}
\toprule
Method \\Name & Macro-P & Macro-R & Macro-F &Micro-F&MRR  \\ \midrule
 Global \\statistics&7$\times10^{-3}$& $10^{-2}$  & 6$\times10^{-3}$ & 0.17 & 0.28 \\\midrule
 Character \\distributions&0.14  &0.10  &0.10 &0.38 &0.48  \\\midrule
 Character \\embeddings&0.38  &0.33  &0.33 &0.54 &0.64  \\\midrule
 Word\\ embeddings&0.18  &0.16 &0.15 &0.47 & 0.58 \\\midrule
 Paragraph\\ embeddings&0.23  & 0.19 & 0.19&0.43 & 0.54 \\\midrule
 All features&0.46  & 0.40 &0.40 &0.62 &0.71  \\\midrule
\midrule
BERT with \\only values&0.46  & 0.45 &0.43 & 0.64&0.73  \\\midrule
SeLaB&\textbf{0.55}  &\textbf{0.53}  &\textbf{0.51} &\textbf{0.72} &\textbf{0.80}  
 \\ \bottomrule
\end{tabular}
\end{table}
\fi
Figure \ref{accuracy}(a) shows the top-$k$ accuracy results where our method
SeLaB outperforms the baselines. BERT with only values, Sherlock, and all features baselines have close performance. So, the BERT-based embedding, which is trained by using only data values, is as good as the hand-crafted features. While extracting the hand-crafted features requires significant human effort to compute the global statistics, character distributions and three types of embeddings (character, word, and paragraph), BERT embeddings are trained jointly with the classification layer with minimal preprocessing which reduces the human effort.

\iffalse
\begin{figure}[!t]
\centerline {\includegraphics[width=3.7in]{wikitables_accuracy.png}}
\vspace{-.2in}
\vskip +0.1 true in \caption{Top-k accuracy results for WikiTables collection}
\label{wikitables_accuracy}
\end{figure}
\fi

Table \ref{metrics}(b) and Figure \ref{accuracy}(b) show the performance of different approaches on the Log tables from network equipment. Consistent with WikiTables, our results on the log tables corpus 
show the importance of a column's context in improving the semantic labels prediction, especially for top-1 accuracy as shown in Figure \ref{accuracy}(b). The top-5 accuracies for SeLaB, BERT with only values, and all features baseline are similar which indicates that the ambiguity in semantic labeling occurs mainly when predicting exact schema labels. Semantic labeling results on WikiTables and Log tables show that SeLaB achieves significant improvements in the evaluation metrics of two data table collections from different domains, which supports the generalization characteristic of our proposed method.

\iffalse
\begin{table}[]
\centering
\caption{Logs tables schema labels prediction}
\label{log_results}
\begin{tabular}{@{}llllll@{}}
\toprule
Method \\Name & Macro-P & Macro-R & Macro-F &Micro-F&MRR  \\ \midrule
 Global \\statistics& 0.16  & 0.16  & 0.15 & 0.54 & 0.64 \\\midrule
 Character \\distributions&0.40  &0.42  &0.40 &0.68 & 0.71 \\\midrule
 Character \\embeddings&0.44 &0.44  &0.43 &0.69 & 0.71 \\\midrule
 Word\\ embeddings&0.35  &0.34&0.33 &0.65 & 0.69 \\\midrule
 Paragraph\\ embeddings&0.31  &0.29  &0.29 &0.61 & 0.67 \\\midrule
 All features&0.50  &0.49  &0.49 &0.71 & 0.72 \\\midrule
\midrule
BERT with \\only values&0.48  &0.49  &0.47 &0.73 & 0.76 \\\midrule
SeLaB&\textbf{0.62}  &\textbf{0.63}  &\textbf{0.61} &\textbf{0.81} &  \textbf{0.81}
 \\ \bottomrule
\end{tabular}
\end{table}
\fi

\iffalse
\begin{figure}[!t]
\centerline {\includegraphics[width=3.7in]{}}
\vspace{-.2in}
\vskip +0.1 true in \caption{Top-k accuracy results for logs tables collection}
\label{log_accuracy}
\end{figure}
\fi

\iffalse
\begin{figure}[ht]
    \centering
     \includegraphics[width=0.23\textwidth]{alice.accuracy.vs.freq.png}
    \includegraphics[width=0.23\textwidth]{wiki.accuracy.vs.freq.png} 
    \vspace*{-4mm}
    \caption{Accuracy vs. Number of Training Samples}
    \label{fig:accuracy.freq}
\end{figure}
\fi

\begin{table*}[!h]
\centering
\resizebox{\textwidth}{!}{\begin{tabular}{@{}lll@{}}
\toprule
Table attributes & First phase predictions & Third phase predictions  \\ \midrule
 season, level, division, section, position & year, level, division, division, finish  & \textbf{season}, level, division, \textbf{section}, \textbf{position}  \\
 year, title, developer, publisher, setting, platform & year, title, developer, developer, setting, system  & year, title, developer, \textbf{publisher}, setting, system  \\
 pos, rider, bike, pts & rank, rider, bike, pts  & \textbf{pos}, rider, bike, pts  \\
 pos, class, no, team, drivers, car, laps, qual pos & pos, group, rank, driver, driver, car, deaths, rank  & pos, \textbf{class}, \textbf{no}, entrant, \textbf{drivers}, car, \textbf{laps}, grid  \\
 senator, party, years, term, electoral history& representative, party, years, wins, electoral history & representative, party, years, \textbf{term}, electoral history \\
 date, time, home team, away team& date, time, home team, home team & date, time, home team, \textbf{away team} \\
 site, location, year, description& site, province, year, description&name, \textbf{location}, year, description \\
 land area, latitude, longitude& area, latitude, geographic coordinate system & \textbf{land area}, latitude, \textbf{longitude} \\
 title, director, cast, genre, notes&film, role, cast, genre, notes &role, \textbf{director}, cast, genre, notes\\
 county, location, exit number, destinations, notes&county, location, notes, notes, notes &county, location, exit, \textbf{destinations}, notes
 \\ \bottomrule
\end{tabular}}
\vspace*{+1mm}
\caption{Example of predicted schema labels from Wikitables testing set}
\label{examples}
\end{table*}

\begin{figure}[!t]
\centerline {\includegraphics[width=3in]{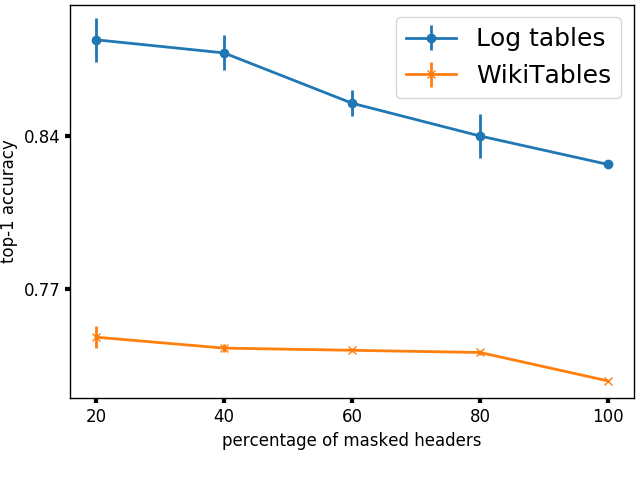}}
\vspace{-.2in}
\vskip +0.1 true in \caption{SeLaB top-1 accuracy for masked headers}
\label{maked_headers}
\end{figure}

\begin{figure}[!t]
\centerline {\includegraphics[width=3in]{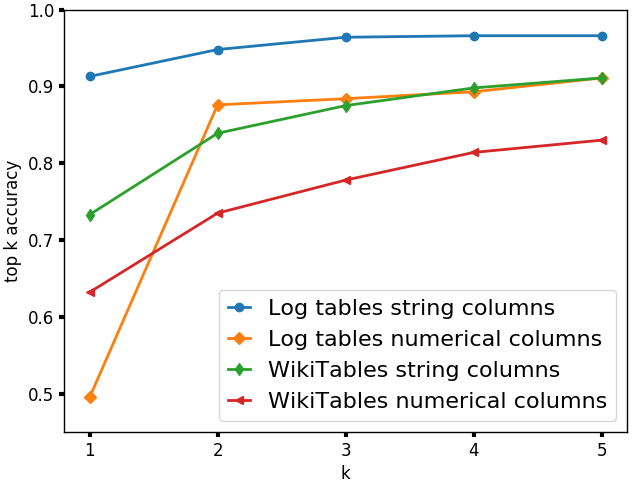}}
%\vspace*{-6mm} 2.5
\vskip +0.1 true in \caption{Top-k accuracy for string and numerical columns}
\label{string_numerical}
\end{figure}

\subsubsection{Masked headers}
We showed the performance of SeLaB where all semantic
labels are missing for a given data 
table. This can be seen as an extreme case. A common scenario for tables extraction is to have a percentage of missing or false headers. To better understand how SeLab deals with such tables, we randomly mask a percentage of headers from tables in the testing set, and we report the top-1 accuracy function of the percentage of masked headers as shown in Figure \ref{maked_headers}. For each table, we run the masked headers for five times with a randomly selected headers, and we compute mean and  standard deviation (std) for each percentage of masked headers. $100\%$ masked headers means that all labels are missing which is the most difficult and the main setting for SeLaB. 

As reported in Figure \ref{maked_headers}, the maximum std (vertical bar) is $0.01$ for Log tables and $0.005$ for WikiTables. Figure \ref{maked_headers} shows that when the percentage of masked headers decreases, there is an increase in the mean of top-1 accuracy for predicted masked headers. This means that the attribute's context is more accurate given that the labels of the non-masked headers are groundtruth labels. However, comparing the fully predicted context in $100\%$ masked headers with the partially predicted context in $20\%$ masked headers, there is only a small improvement (mainly for WikiTables) which indicates that the predicted context in the extreme case is as good as groundtruth context.

\subsubsection{String vs Numeric columns analysis}
We evaluate the performance of SeLaB for two categories of columns which are numeric and string. As shown in Figure \ref{string_numerical}, we report top-$k$ accuracy of numerical and string columns for Log tables and WikiTables. In both datasets, semantic labeling of string columns outperforms numerical columns. So, generating an exact schema label for numerical data values is more ambiguous than string values, as numerical columns contain similar values.

%\begin{figure}[ht]
%    \centering
%     \includegraphics[width=0.23\textwidth]{alice.accuracy.vs.freq.png}
%    \includegraphics[width=0.23\textwidth]{wiki.accuracy.vs.freq.png} 
%    \vspace*{-4mm}
%    \caption{Accuracy vs. Number of Training Samples}
%    \label{fig:accuracy.freq}
%\end{figure}

\begin{figure*}[t!]
    \centering
    \begin{subfigure}[t]{0.5\textwidth}
        \centering
        \includegraphics[height=3.5in]{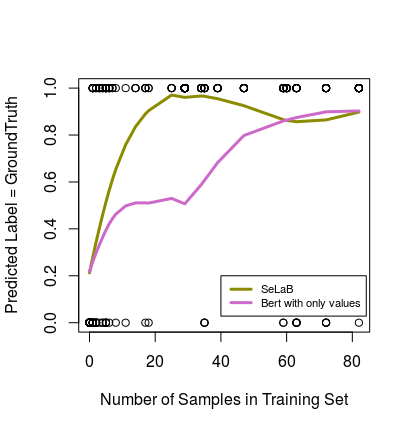}
        %\vspace*{-4mm} 0.18 1.8
        \caption{Log tables}
        
    \end{subfigure}%
    ~ 
    \begin{subfigure}[t]{0.5\textwidth}
        \centering
        \includegraphics[height=3.5in]{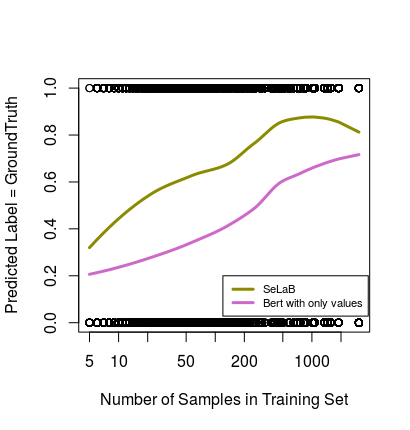}
        %\vspace*{-4mm} .035 1.8
        \caption{WikiTables}
    \end{subfigure}
    %\vspace*{-4mm}
    \caption{Accuracy vs. Number of training samples per label}
    \label{fig:accuracy.freq}
\end{figure*}

\subsubsection{Predicted labels examples}
To better understand how SeLaB works, we show examples of predicted schema labels from WikiTables testing set in Table \ref{examples}. Each row corresponds to a testing table where we show the ground truth attributes, first phase predictions, and the final predicted schema labels. For example, for the first row, there are three wrong predictions (\textit{year} instead of \textit{season}, \textit{division} instead of \textit{section}, and \textit{finish} instead of \textit{position}) from first phase predictions which are based only on data values. After incorporating context for each attribute, SeLaB updates the predicted label for each column, and the new context-aware semantic labels that match the ground truth labels are shown in bold in Table \ref{examples}. For the first example, we obtain the third phase predictions which are identical to the ground truth labels after three corrections from context-aware representation for each column. %The correctly updated labels in the third phase, comparing to the first phase, are shown in bold in Table \ref{examples}.
For the sixth row, the table's attributes contain \textit{home team} and \textit{away team}. Both attributes are predicted \textit{home team} after first phase predictions. SeLaB learns that \textit{away team} occurs with \textit{home team} so that the predictions are corrected after the third phase.

\subsubsection{Effect of the Number of Training Samples}
To understand how the number of training
samples for each semantic label affects the accuracy of SeLaB predictions 
in the test data, in  Figure \ref{fig:accuracy.freq}, for each label in the testing set, we plot the indicator values of correct SeLaB prediction against the number of samples for each label in the training set as black circles 
(i.e. 1 indicates the 
predicted column header is the same as the ground truth). Local smoothing \cite{cleveland.loess} was performed to obtain the average accuracy curve for SeLaB predictions (yellow line). As a reference, we also added a similar local smoothing curve  representing the accuracy curve from predictions obtained using ``Bert with only values'' (pink line). 

Figure \ref{fig:accuracy.freq} clearly demonstrates that overall speaking, as the number of samples for each label in the training set increases, both SeLaB and ``Bert with only values'' are performing better. However, SeLaB
appears to perform much better when there is sufficient number of samples per label (e.g. more than 20 instances). There is a slight dip for the SeLab curve for those columns when the corresponding number of samples in the training set becomes the largest. Upon close inspection, it does not seem to imply that the performance of SeLaB is deteriorating. For example, in Wikitables, the column header with the largest number of instances is in fact a generic label \textit{name} (with 3064 instances in training data). In the testing set, SeLaB sometimes predicts \textit{name} as \textit{player}, \textit{swimmer}, \textit{representative}, etc,  depending on the context of the table, thus potentially yielding more appropriate header names. %--
 %labels.

\iffalse
\begin{figure}[!t]
\centerline {\includegraphics[width=3in]{masked_headers.png}}
\vspace{-.2in}
\vskip +0.1 true in \caption{Top-1 accuracy results for masked headers}
\label{wikitables_accuracy}
\end{figure}
\fi

\section{Conclusions}

We have shown that a context-aware model that combines data values and column's context outperforms approaches that predict semantic labels only on the basis of data values. Our method has been evaluated on two real-world datasets from multiple domains: WikiTables extracted from Wikipedia and Log tables from network equipment. We have shown that the attribute's predicted context, which is incorporated into our model SeLaB, solves the ambiguity in semantic labels predictions. Our model is trained end-to-end for both feature extraction and label prediction which reduces the human effort in semantic labeling.

%the schema label generation process.
Future work includes looking at how to incorporate metadata of each data table, such as table caption and description, into SeLaB,  % to reduce the time complexity of our method, 
and how to select the best subset of data values for each column to improve semantic labeling results.		

\bibliographystyle{ACM-Reference-Format}
\bibliography{sigconf}
\end{document}